\documentclass[runningheads]{llncs}
\usepackage[T1]{fontenc}
\usepackage{graphicx}
\usepackage{microtype}
\usepackage{hyperref}
\usepackage{url}
\usepackage{booktabs}
\usepackage{graphicx}
\usepackage[utf8]{inputenc}
\usepackage[T1]{fontenc}
\usepackage{amssymb,amsmath}
\usepackage[most]{tcolorbox}
\usepackage{xcolor}
\usepackage{listings}
\usepackage{float}
\usepackage{subcaption}
\usepackage{bbm}
\usepackage{indentfirst}
\usepackage{microtype}
\usepackage{lineno}
\usepackage{needspace}
\usepackage{caption}

\usepackage{fvextra}

\DefineVerbatimEnvironment{leanverbatim}{Verbatim}{
  fontsize=\small,
  breaklines=true,
  breakanywhere=true
}

\definecolor{darkblue}{rgb}{0, 0, 0.5}
\definecolor{leanbg}{RGB}{245,248,248}
\definecolor{leanborder}{RGB}{180,200,210}
\definecolor{titlebg}{RGB}{75,150,190}

\newtcolorbox{examplebox}[2][]{%
  enhanced,
  colback=white,
  colframe=leanborder,
  boxrule=0.8pt,
  arc=3pt,
  left=8pt,right=8pt,top=10pt,bottom=8pt,
  title={\parbox[t]{\linewidth}{\strut #2}},
  coltitle=white,
  colbacktitle=titlebg,
  fonttitle=\bfseries,
  varwidth boxed title*=-1.2cm
  attach boxed title to top left={xshift=0mm,yshift=-2mm},
  boxed title style={
    arc=3pt,
    outer arc=3pt,
    boxrule=0pt,
    left=8pt,right=8pt,top=3pt,bottom=3pt
  },
  #1
}

\newtcolorbox{codepanel}{%
  enhanced,
  colback=leanbg,
  colframe=leanborder,
  boxrule=0.6pt,
  arc=2pt,
  left=6pt,right=6pt,top=6pt,bottom=6pt,
}

\newcommand{\panelheading}[1]{\vspace{2pt}\noindent\textbf{#1}\par\vspace{4pt}}

\DeclareUnicodeCharacter{2115}{\ensuremath{\mathbb{N}}} 
\DeclareUnicodeCharacter{211D}{\ensuremath{\mathbb{R}}} 
\DeclareUnicodeCharacter{2200}{\ensuremath{\forall}}    
\DeclareUnicodeCharacter{2192}{\ensuremath{\to}}        
\DeclareUnicodeCharacter{2264}{\ensuremath{\le}}        
\DeclareUnicodeCharacter{2223}{\ensuremath{\mid}}       
\DeclareUnicodeCharacter{03B1}{\ensuremath{\alpha}}     
\DeclareUnicodeCharacter{03B2}{\ensuremath{\beta}}      

\definecolor{premisehl}{RGB}{255,236,160}

\newcommand{\premhl}[1]{%
  \colorbox{premisehl}{\strut\ttfamily\small #1}%
}

\DefineVerbatimEnvironment{leanhlverbatim}{Verbatim}{
  fontsize=\small,
  breaklines=true,
  breakanywhere=true,
  commandchars=\\\{\}
}

\begin{document}
\title{TheoremBench: Evaluating LLMs on Theorem Proving in Formal Mathematics}
\titlerunning{TheoremBench: LLMs in Formal Mathematics}
%
\author{QuocViet Pham\inst{1, 2}\orcidID{0009-0005-5880-9608} \and
Elvir Karimov\inst{1,4}\orcidID{0009-0002-3198-6186} \and
Andrey Galichin\inst{1,3,4}\orcidID{0009-0002-5918-1435} \and Ivan Oseledets\inst{1,3}\orcidID{0000-0003-2071-2163}\\} 
\authorrunning{Q. Pham et al.}
%
\institute{Skolkovo Institute of Science and Technology \and
HSE University \and Artificial Intelligence Research Institute \and Sberbank\\
\email{\{QuocViet.Pham, elvir.karimov\}@skoltech.ru, a.v.galichin@mtuci.ru, ivan.oseledets@gmail.com}}
\maketitle              
\begin{abstract}
LLMs have recently achieved strong results on formal proving benchmarks. However, existing evaluations remain heavily concentrated on competition-style problems and often fail to capture how models behave on longer, more dependency-rich mathematical developments. We introduce \textbf{TheoremBench}, a Lean4 benchmark designed to evaluate theorem provers beyond contest settings. The benchmark is built from nearly one hundred classical theorems and is released in two complementary forms: a \emph{plain main} version containing one target theorem per instance, and a \emph{premised} version that expands each theorem into a structured family of related proving tasks consisting of the main theorem together with automatically extracted supporting subtheorems. This design enables evaluation of not only whether the final theorem was proved from scratch, but also of partial progress through the internal proof structure of a theorem. Our experiments show that explicit premises substantially improve performance for Lean4-capable prover models. To provide a comprehensive evaluation, we introduce theorem-level coverage and token-efficiency metrics that expose qualitative differences in proof behavior. The results show that current provers remain strongly biased toward easy subtheorems and often solve theorems through long and inefficient tactic traces rather than compact proof plans. TheoremBench therefore provides a more fine-grained view of formal reasoning ability and highlights the importance of structural benchmark design for evaluating Lean4 theorem provers.

\end{abstract}
\section{Introduction}
Formal proof assistants such as Lean4 encode mathematical statements and proofs so that every inference is checked by a trusted kernel~\cite{Lean4OriginalPaper}. With libraries such as mathlib~\cite{Lean4Mathlib}, they support large-scale, reusable formal mathematics. This makes Lean theorem proving a useful testbed for LLM reasoning: unlike natural-language answers, generated proofs are accepted or rejected by a verifier.

LLM-based provers have advanced rapidly on competition-style benchmarks. Kimina-Prover~\cite{kimina_prover_2025}, DeepSeek-Prover-V2k~\cite{ren2025deepseekproverv2advancingformalmathematical}, Goedel-Prover~\cite{lin2025goedelproverv2scalingformaltheorem}, and related ATP-augmented systems combine language modeling, search, reinforcement learning, and verifier feedback. Yet Lean proving remains difficult: a model must generate valid code, instantiate implicit arguments, select mathlib lemmas, respect namespaces and notation, and satisfy the type checker.

Benchmarks such as MiniF2F~\cite{zheng2022minif2fcrosssystembenchmarkformal}, PutnamBench~\cite{tsoukalas2024putnambenchevaluatingneuraltheoremprovers}, ProverBench~\cite{ren2025deepseekproverv2advancingformalmathematical}, CombiBench~\cite{liu2025combibenchbenchmarkingllmcapability}, and FormalMATH~\cite{yu2025formalmathbenchmarkingformalmathematical} have driven progress, but mainly evaluate standalone contest-style statements. Real Lean developments are more structured: final theorems depend on intermediate lemmas, local definitions, typeclass assumptions, namespaces, and prior results. Thus, final-statement evaluation often hides whether a model can use or reconstruct the surrounding proof structure.

We introduce TheoremBench, a Lean4 benchmark for this intermediate regime. It is based on classical mathematical theorems inspired by Wiedijk's list of 100 theorems~\cite{FreekWiedijk} and has two forms. The \emph{plain-main} dataset presents each target theorem as a standalone final statement. The \emph{premised} dataset expands each theorem into related Lean instances: the main theorem and automatically extracted subtheorems, with relevant prior results exposed as explicit premise binders. This supports evaluation of both final success and partial progress through the proof structure.

\begin{figure}[t]
\includegraphics[width=0.9\linewidth]{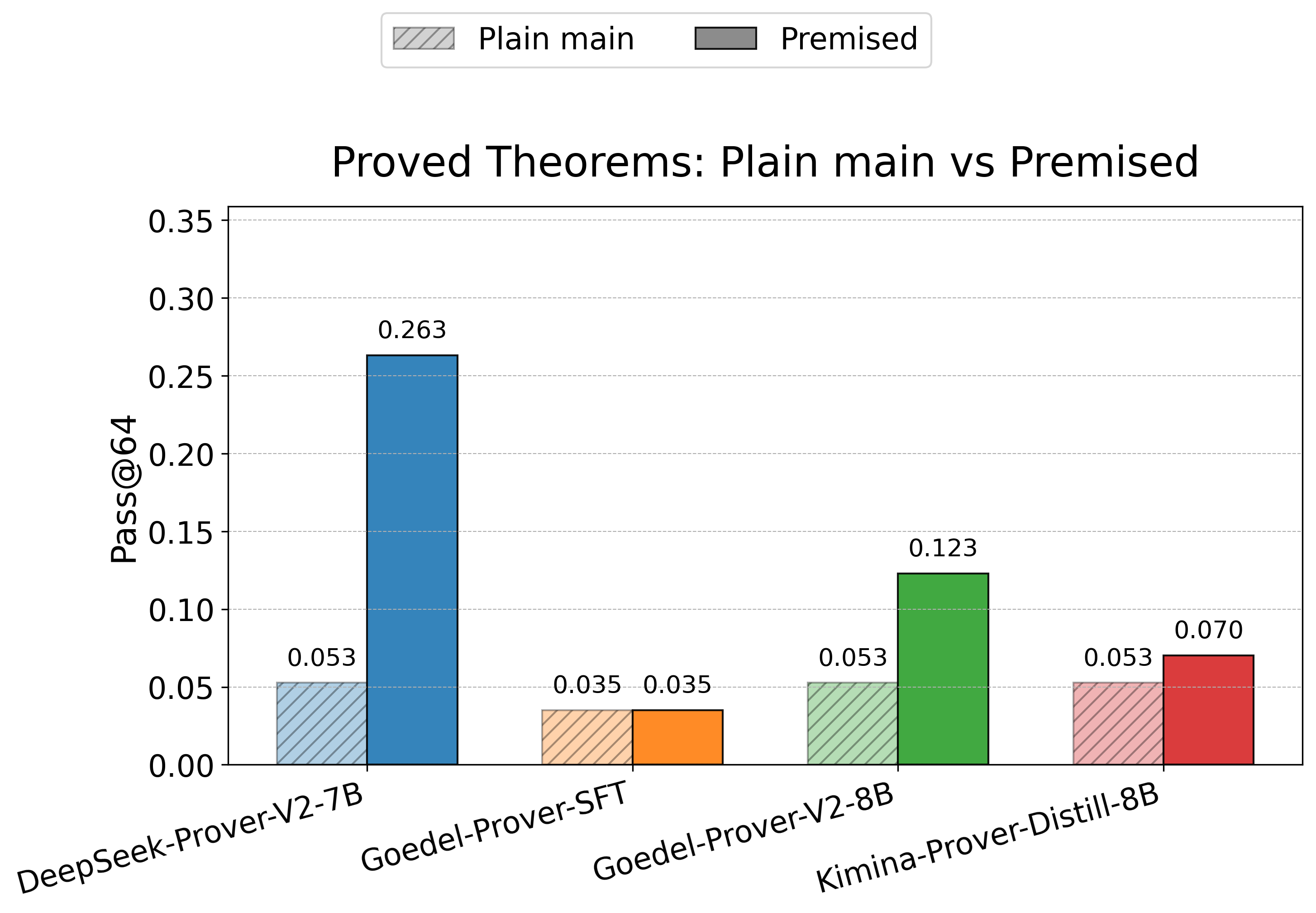} 
\caption{Performance comparison of Lean4-capable theorem provers in the plain main and premised settings, indicating their theorem fully-proved ability. Premises help substantially for DeepSeek and Goedel-Prover-V2-8B, modestly for Kimina, and nothing for non-reasoning model Goedel-Prover-SFT.}
\label{fig:plain_vs_premised_pass64}
\end{figure}

A final theorem statement often lacks the context needed to recover supporting lemmas, while accumulating lemmas into long prompts scales poorly. TheoremBench instead embeds selected dependencies directly into theorem declarations as explicit assumptions, producing self-contained Lean snippets. This transformation must preserve imports, namespaces, implicit binders, typeclass inference, notation, and local context, yielding a controlled and diagnostic evaluation setting.

Our experiments show that explicit premises reduce difficulty only for models able to exploit them. DeepSeek-Prover-V2-7B and Goedel-Prover-V2-8B benefit strongly, Kimina-Prover-Distill-8B improves modestly, and Goedel-Prover-SFT shows almost no gain. The premised structure also enables theorem-level analysis: we measure how many supporting subtheorems are proved and how close a model comes to completing a full theorem group.

\begin{figure}[t]
\includegraphics[width=0.9\linewidth]{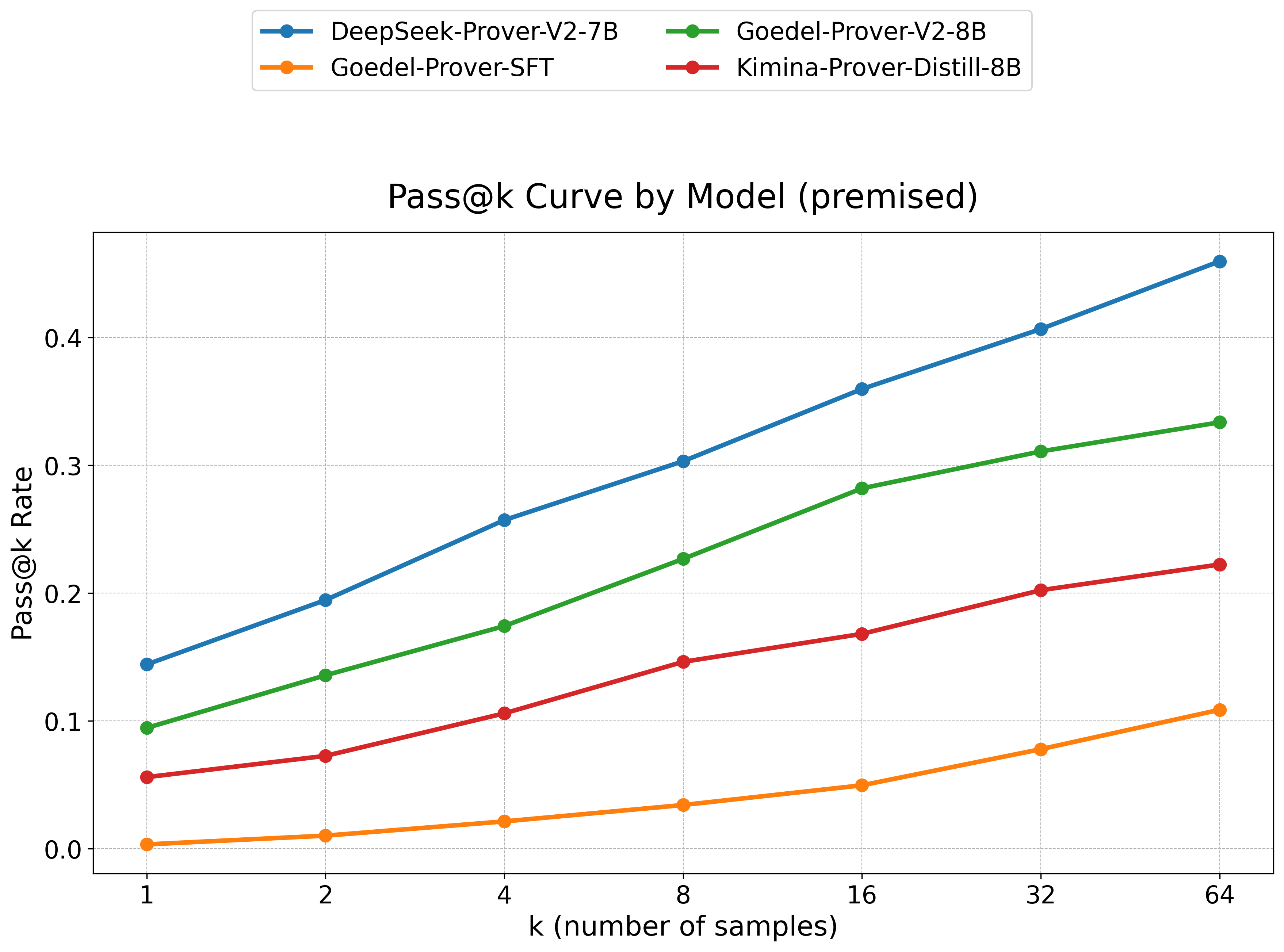}
\caption{Performance of models on the premised benchmark. DeepSeek-Prover-V2-7B consistently achieves the highest success rate, followed by Goedel-Prover-V2-8B, Kimina-Prover-Distill-8B, and Goedel-Prover-SFT}
\label{fig:passk_curve_all_models}
\end{figure}

Our contributions are:
\begin{itemize}
    \item We introduce TheoremBench, a Lean4 benchmark for mid-level classical theorems beyond competition-style datasets.
    \item We propose a premised representation that converts relevant prior Lean results into explicit premise binders, producing self-contained dependency-aware proving instances.
    \item We add theorem-level coverage and token-efficiency metrics, showing differences in premise use, proof-structure coverage, and proof compactness beyond pass@k.
\end{itemize}

\section{Related Work}

\paragraph{Autoformalization and proof assistants.}
Autoformalization maps informal mathematical statements and proofs to verifier-checkable formal code. Early LLM-based work showed that informal competition problems can be translated into formal statements and used to improve proving performance \cite{wu2022autoformalizationlargelanguagemodels}, while ProofNet connected informal and formal undergraduate mathematics through aligned statements and proofs \cite{azerbayev2023proofnetautoformalizingformallyproving}. Surveys cover the broader development of autoformalization, neural theorem proving, premise selection, proof-step generation, search, synthetic data, and evaluation protocols \cite{weng2025autoformalizationerasurvey,li2024surveydeeplearningtheoremproving}. A central difficulty is the gap between informal notation and formal syntax: Lean, Isabelle, and Coq require exact binders, types, references, and proof terms \cite{Lean4OriginalPaper,paulson1994isabelle,1997Coq}. Lean4 is especially relevant because it combines a proof assistant with a programming language and is supported by mathlib \cite{avigad2021theoremprovinginlean4,Lean4Mathlib}, but this also exposes models to namespaces, typeclass inference, scoped notation, and evolving APIs \cite{wang2024theoremllamatransforminggeneralpurposellms,ying2025leanworkbooklargescalelean}.

\paragraph{LLM-based formal theorem proving.}
Neural theorem proving predates current LLM provers: GamePad studied Coq proof-state learning \cite{huang2018gamepad}, TacticToe learned tactic search in HOL4 \cite{gauthier2018tactictoe}, HOList and DeepHOL framed higher-order theorem proving as reinforcement learning \cite{bansal2019holist}, and CoqGym provided a large Coq dataset with AST-based tactic generation \cite{yang2019coqgym}. LLM provers extend this line by generating candidate proofs checked by the proof assistant. GPT-f demonstrated accepted formal proof generation \cite{polu2020generativelanguagemodelingautomated}; LeanDojo made Lean interaction and retrieval-augmented proving reproducible, emphasizing premise selection as a mathlib-scale bottleneck \cite{yang2023leandojotheoremprovingretrievalaugmented}. ReProver \cite{yang2023leandojotheoremprovingretrievalaugmented}, REAL-Prover \cite{shen2025realproverretrievalaugmentedlean}, BFS-Prover \cite{xin2025bfsproverscalablebestfirsttree}, and HUNYUANPROVER \cite{li2025hunyuanproverscalabledatasynthesis} combine retrieval, search, synthetic data, and verifier feedback. Other systems target premise selection, search, repair, or decomposition: Thor \cite{jiang2022thor}, Magnushammer \cite{mikula2023magnushammer}, Draft, Sketch, and Prove \cite{jiang2023draftsketchandprove}, HyperTree Proof Search \cite{lample2022hypertree}, Baldur \cite{first2023baldur}, and COPRA \cite{thakur2023copra}. Lean-specific scaling efforts include Llemma \cite{azerbayev2023llemma}, TheoremLlama \cite{wang2024theoremllamatransforminggeneralpurposellms}, Lean Workbook \cite{ying2025leanworkbooklargescalelean}, and LeanNavigator \cite{yin2025leannavigator}.

\paragraph{Benchmarks.}
Formal-proving benchmarks have evolved from large proof-assistant environments to competition and curriculum-style datasets. HOList and CoqGym provide large-scale testbeds for HOL Light and Coq \cite{bansal2019holist,yang2019coqgym}; LeanDojo provides a Lean interaction framework with premise annotations and generalization splits \cite{yang2023leandojotheoremprovingretrievalaugmented}. MiniF2F targets olympiad-style problems \cite{zheng2022minif2fcrosssystembenchmarkformal}, PutnamBench raises the level to undergraduate competition mathematics \cite{tsoukalas2024putnambenchevaluatingneuraltheoremprovers}, ProofNet and FIMO connect informal and formal undergraduate or IMO-style reasoning \cite{azerbayev2023proofnetautoformalizingformallyproving,liu2023fimo}, CombiBench stresses combinatorics \cite{liu2025combibenchbenchmarkingllmcapability}, FormalMATH expands Lean4 evaluation to thousands of problems \cite{yu2025formalmathbenchmarkingformalmathematical}, and ProverBench evaluates modern Lean provers in the DeepSeek-Prover-V2 setting \cite{ren2025deepseekproverv2advancingformalmathematical}. Most of these datasets evaluate standalone theorem statements. TheoremBench instead evaluates classical Lean developments through an explicit premised structure, making supporting subtheorems visible benchmark instances rather than hidden context.

\begin{table}[!htbp]
\begin{center}
\begin{tabular}{llll}
\toprule
\multicolumn{1}{c}{\bf Dataset}  &\multicolumn{1}{c}{\bf Samples}  &\multicolumn{1}{c}{\bf Difficulty}  &\multicolumn{1}{c}{\bf Success} \\
\midrule
miniF2F~\cite{zheng2022minif2fcrosssystembenchmarkformal} &244  &Highschool, AIME            &90.4\%  \\
PutnamBench~\cite{tsoukalas2024putnambenchevaluatingneuraltheoremprovers} &658  &Undergraduate               &13.07\% \\
ProverBench~\cite{ren2025deepseekproverv2advancingformalmathematical} &325  &AIME                        &59.1\%  \\
ProofNet~\cite{azerbayev2023proofnetautoformalizingformallyproving} &186  &Undergraduate               &37.1\%  \\
CombiBench~\cite{liu2025combibenchbenchmarkingllmcapability} &100  &Undergraduate               &12\%    \\
FormalMATH-Lite~\cite{yu2025formalmathbenchmarkingformalmathematical} &425  &Highschool, Undergraduate   &54.11\% \\
FormalMATH-All~\cite{yu2025formalmathbenchmarkingformalmathematical} &5560 &Highschool, Undergraduate   &-       \\
\hline
TheoremBench (Ours) &1142 &Undergraduate, Postgraduate &54.91\% \\
\bottomrule
\end{tabular}
\end{center}
\caption{Current popular benchmarks and reported success of strong prover models.}
\label{sample-table}
\end{table}

\paragraph{Evaluation metrics and efficiency.}
LLM-based formal proving is usually evaluated by pass@k: a theorem is solved if one of the first $k$ generated attempts is accepted by the verifier. This follows code-generation evaluation \cite{chen2021evaluatingcodellms} and is standard in MiniF2F \cite{zheng2022minif2fcrosssystembenchmarkformal}, LeanDojo \cite{yang2023leandojotheoremprovingretrievalaugmented}, DeepSeek-Prover \cite{ren2025deepseekproverv2advancingformalmathematical}, Kimina-Prover \cite{kimina_prover_2025}, Goedel-Prover \cite{lin2025goedelproverfrontiermodelopensource}, and FormalMATH \cite{yu2025formalmathbenchmarkingformalmathematical}. Since pass@k ignores efficiency, search-based provers also report explored states, verifier calls, CPU time, and depth, as in HyperTree Proof Search \cite{lample2022hypertree}, DeepSeek-Prover-V1.5 \cite{xin2025deepseekprover15}, BFS-Prover \cite{xin2025bfsproverscalablebestfirsttree}, and APOLLO \cite{ospanov2025apollo}. Proof quality adds another axis: Baldur and COPRA study repair and interactive/backtracking behavior \cite{first2023baldur,thakur2023copra}, InternLM2.5-StepProver reports proof length and CPU usage \cite{wu2024internlmstepprover}, and ProofOptimizer trains models to shorten Lean proofs \cite{gu2025proofoptimizer}. TheoremBench follows this diagnostic direction by retaining pass@k while adding theorem-level coverage, fully proved parent counts, and token-efficiency.

\section{TheoremBench Dataset Construction}

TheoremBench is a Lean4 benchmark for evaluating theorem provers on classical mathematical theorems beyond competition-style tasks. The benchmark is constructed from Lean4 formalizations of classical results inspired by Freek Wiedijk's list of 100 theorems~\cite{FreekWiedijk}. Unlike benchmarks centered on compact olympiad-style statements, these formalizations usually appear inside larger Lean developments and depend on imported theories, local definitions, namespaces, typeclass assumptions, and intermediate lemmas.

\subsection{Construction Pipeline}
For each source file, we construct the dataset as follows:

\begin{enumerate}
\item We collect raw Lean4 theorem developments from classical formalizations inspired by Wiedijk's list of 100 theorems~\cite{FreekWiedijk}. Each source file corresponds to one parent theorem and contains the final theorem together with its surrounding Lean development.

\item For each parent theorem, we parse the Lean file into a theorem group. The final declaration is treated as the main theorem, and the required Lean context is reconstructed so that the extracted instance can compile independently. Each parent theorem also provides supporting theorem and lemma declarations, which we extract as subtheorems and use to build the premised dataset. 

\item For every theorem or subtheorem, earlier results used in its proof are converted into explicit premise binders. We then emit a self-contained Lean4 snippet, verify it with Lean4, keep only compilable instances, and store each instance with its theorem-level ID, Lean declaration name, extracted premises, environment context, and ground-truth proof.
\end{enumerate}

\begin{figure}[t]
    \centering
    \includegraphics[width=0.99\linewidth]{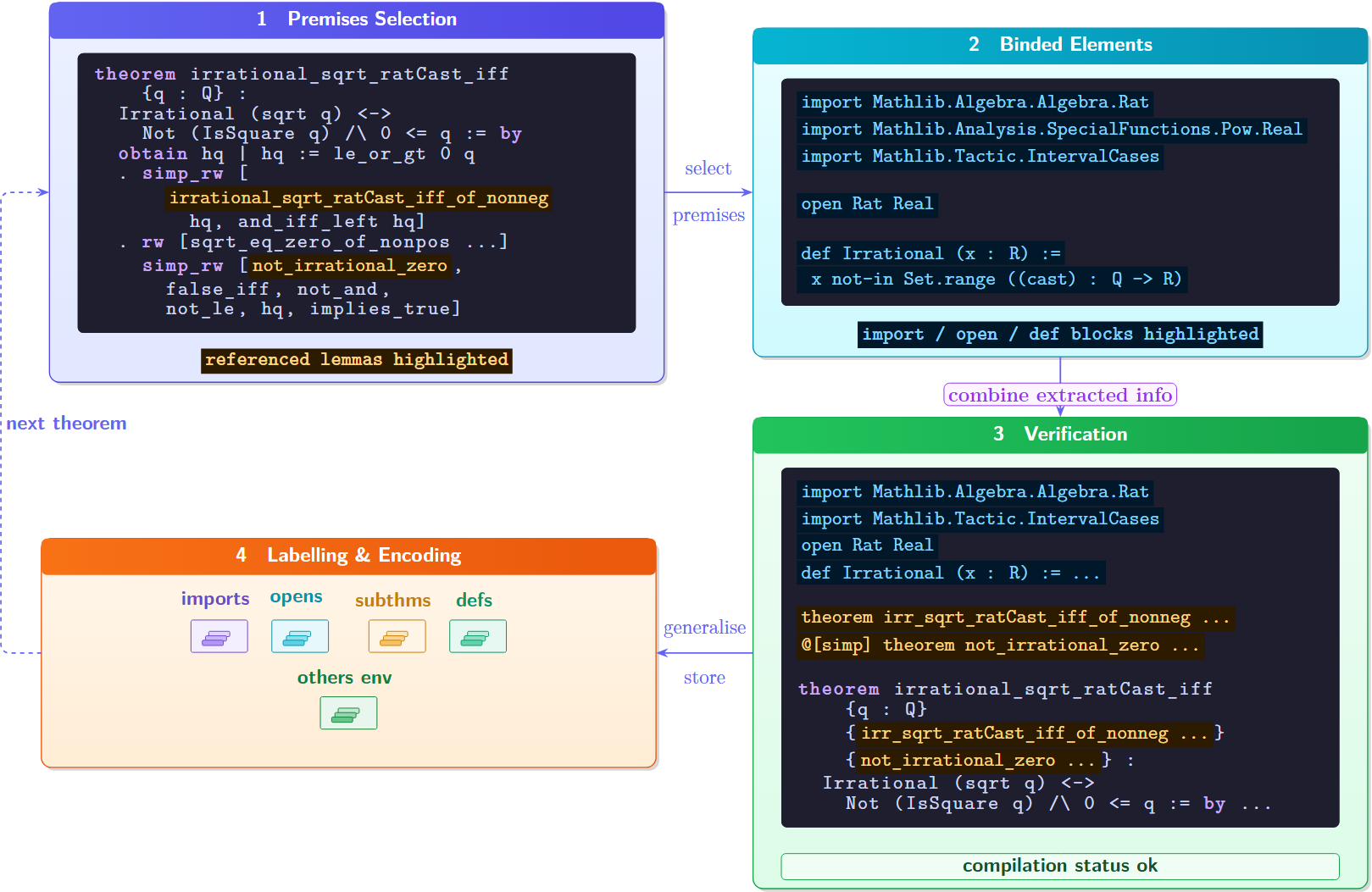}
    \caption{TheoremBench construction pipeline. Lean4 source files are parsed into theorem groups, enriched with required formal context, transformed into plain-main and premised instances, and verified by Lean4 before inclusion in the benchmark.}
    \label{fig:collection_pipeline}
\end{figure}

This procedure produces two aligned dataset views. The \emph{plain-main} version contains one standalone final theorem per parent theorem. The \emph{premised} version expands the same parent theorem into a group of related proving tasks, consisting of the main theorem and its extracted supporting subtheorems. Figure~\ref{fig:collection_pipeline} illustrates the full construction pipeline.

\subsection{Dataset Statistics}
TheoremBench covers a broad range of undergraduate and postgraduate mathematics, including algebra, number theory, real and complex analysis, geometry, topology, set theory, logic, combinatorics, probability, and abstract algebra. The difficulty of instances also varies substantially: some require short local reasoning, while others depend on many auxiliary lemmas, definitions, tactics, and explicit binders from mathlib or from earlier results in the same source file.

\begin{figure}[t]
    \centering
    \includegraphics[width=1.0\linewidth]{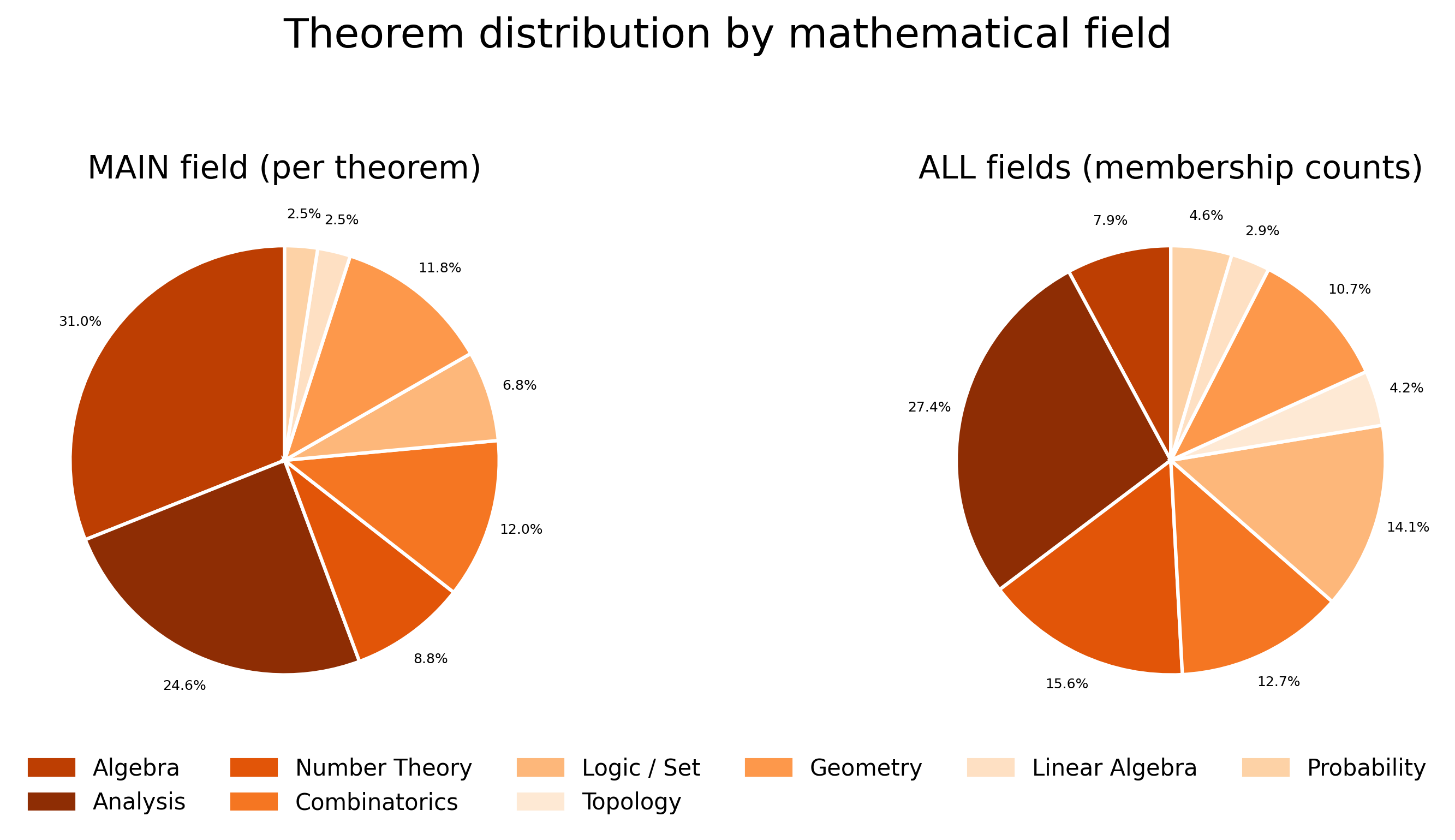}
    \caption{Distribution of TheoremBench instances by mathematical field.}
    \label{fig:pie_fields_count}
\end{figure}

We estimate difficulty using both LLM-based classification and Lean4-level structural features, including proof length, referenced declarations, tactic usage, and the number of mathematical fields involved. Figure~\ref{fig:pie_fields_count} summarizes the field distribution of the benchmark.

\section{Experiments and Discussion}
\subsection{Evaluation metrics}
We evaluate model performance using both standard instance-level metrics and additional theorem-level metrics that are specific to the premised structure of TheoremBench.

\paragraph{Pass@k.}
For a theorem instance, let $r_1,\dots,r_k$ denote the first $k$ generated proof attempts, and let $s_i \in \{0,1\}$ indicate whether attempt $i$ is accepted by Lean4. We define
\begin{equation}
\mathrm{Pass@}k = \mathbbm{1}\!\left(\bigvee_{i=1}^k s_i = 1\right).
\end{equation}
At the dataset level, pass@k is the mean of this indicator over all theorem instances. We report pass@k for powers of two up to $k_{\max}=64$, which reveals both final success and sample-efficiency scaling.

\paragraph{Fully proved parent theorem.}
In the premised dataset, a parent theorem is considered \emph{fully proved} if and only if the model proves the main theorem together with all of its associated subtheorems. This metric is substantially stricter than instance-level pass@k, because a parent theorem is counted as solved only when the entire proof structure has been completed.

\paragraph{Theorem-level coverage.}
For each parent theorem $T$, let $m(T)$ be the total number of subtheorems in its theorem group and let $p(T)$ be the number proved successfully by the model. We define theorem-level coverage by
\begin{equation}
\mathrm{Cov}(T)=\frac{p(T)}{m(T)} \times 100.
\end{equation}
This allows us to study partial progress even when the full theorem group is not solved. In addition to average coverage, we report a survival-style curve showing the fraction of parent theorems whose coverage is at least a chosen threshold.

\paragraph{Average subtheorems proved per theorem.}
To provide a direct interpretation of model progress within a decomposed theorem, we also compute the average number of proved subtheorems per parent theorem:
\begin{equation}
\mathrm{AvgSub} = \frac{1}{N}\sum_{T} p(T),
\end{equation}
where $N$ is the number of parent theorems. This metric answers the practical question: if a theorem is decomposed into intermediate obligations, how many of them will the model solve on average?

\paragraph{Token-efficiency.}
For every solved subtheorem instance, we compare the token length of successful generated proofs against the token length of the ground-truth Lean proof. Let $\ell_{\mathrm{gt}}(T)$ denote the token count of the reference proof, and let $\ell_{\mathrm{gen},1}(T),\dots,\ell_{\mathrm{gen},m}(T)$ denote the lengths of all successful generated proofs among the sampled attempts. We define
\begin{equation}
E_{\mathrm{avg}}(T)=\frac{\frac{1}{m}\sum_{j=1}^m \ell_{\mathrm{gen},j}(T)}{\ell_{\mathrm{gt}}(T)},
\qquad
E_{\min}(T)=\frac{\min_j \ell_{\mathrm{gen},j}(T)}{\ell_{\mathrm{gt}}(T)}.
\end{equation}
Before tokenization, we remove non-Lean explanatory prose and comments from generated outputs, so the metric reflects proof text rather than natural-language reasoning. Values above $1$ indicate that the model tends to produce longer proofs than the reference proof, while values near or below $1$ indicate more compact outputs.

\paragraph{Plain-main versus premised comparison.}
Since the plain-main and premised datasets represent the same underlying theorem families at different structural levels, we also compare pass@64 across the two settings on the theorem units that are shared between them. This comparison measures whether explicit premises help models solve the final theorem more effectively.
\subsection{Evaluating Formal Theorem Provers}

We evaluate four Lean4-capable provers on two aligned versions of TheoremBench: DeepSeek-Prover-V2-7B\cite{ren2025deepseekproverv2advancingformalmathematical}, Goedel-Prover-V2-8B\cite{lin2025goedelproverv2scalingformaltheorem}, Kimina-Prover-Distill-8B\cite{kimina_prover_2025}, and the non-reasoning Goedel-Prover-SFT model\cite{lin2025goedelproverfrontiermodelopensource} as baseline.

The \emph{plain-main} benchmark contains one standalone target theorem per parent theorem. The \emph{premised} benchmark expands the same parent theorem into a group of related instances, including the main theorem and automatically extracted supporting subtheorems. All instances in a group share the same theorem-level ID, while each retains its own Lean4 declaration name.

This setup lets us compare final-statement proving with structured proof completion. It tests whether explicit premises reduce difficulty and whether models can complete a theorem's internal proof structure rather than only isolated statements.

All instances are verified with Lean4 using an automated pipeline that records compilation outcomes, resource usage, and structural statistics. We also keep ground-truth proofs for token-level comparison with generated proofs. Each theorem or subtheorem is sampled up to $k_{\max}=64$ times, and every candidate proof is checked by Lean4. We report both standard instance-level metrics and theorem-level metrics specific to the premised benchmark.
\subsection{Main empirical results}

\paragraph{Subtheorem-level proving performance.}
Figure~\ref{fig:passk_curve_all_models} shows the pass@k curves on the premised benchmark. All models improve monotonically as the generation budget increases, but the ranking remains stable throughout the entire range. DeepSeek-Prover-V2-7B is the strongest model at every value of $k$, reaching approximately $0.46$ pass@64. Goedel-Prover-V2-8B is the second strongest at roughly $0.33$, followed by Kimina-Prover-Distill-8B at roughly $0.22$, while Goedel-Prover-SFT remains substantially lower at about $0.11$. Since the same ranking is already visible at $k=1$ and $k=2$, the gap cannot be explained purely by larger sampling budgets; it reflects a real difference in first-attempt and low-budget proving quality.

\paragraph{Impact of explicit premises.}
Figure~\ref{fig:plain_vs_premised_pass64} compares plain-main and premised theorem proving at pass@64 on the theorem units shared between both dataset versions. The results show that explicit premises reduce benchmark difficulty substantially, but the gain depends strongly on the model. DeepSeek-Prover-V2-7B benefits the most, improving from $0.053$ on plain-main theorems to $0.263$ on premised theorems. Goedel-Prover-V2-8B also improves substantially, from $0.053$ to $0.123$. Kimina-Prover-Distill-8B improves only modestly, from $0.053$ to $0.070$. In contrast, Goedel-Prover-SFT shows essentially no difference, remaining at $0.035$ in both conditions. These results indicate that explicit premises are useful only when the prover is able to exploit the additional structure effectively.

\begin{figure}
    \centering
    \includegraphics[width=0.9\linewidth]{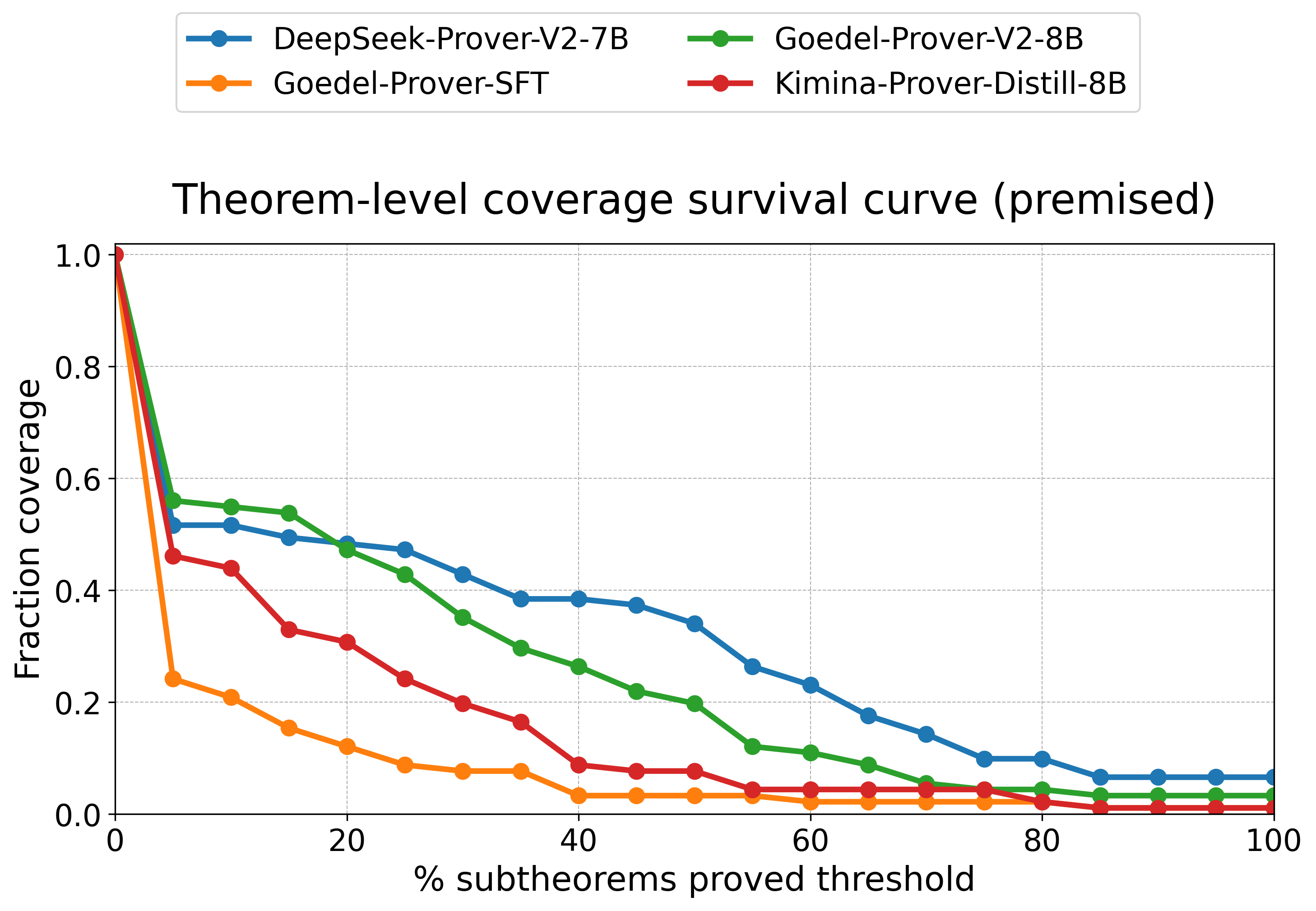}
    \caption{Theorem-level coverage survival curve on the premised benchmark. For each threshold $x$, the y-value gives the fraction of parent theorems for which the model proves at least $x\%$ of subtheorems. Higher curves indicate a greater probability of approaching theorem completion.}
    \label{fig:theorem_coverage_survival}
\end{figure}

\paragraph{Theorem-level completion behavior.}
Instance-level pass@k alone does not show how close a model comes to finishing a full parent theorem. Figure~\ref{fig:theorem_coverage_survival} therefore reports theorem-level coverage in survival form: for each threshold on the x-axis, the y-value gives the fraction of parent theorems for which the model proves at least that percentage of subtheorems. This view is more informative for the premised dataset because it measures partial completion of a theorem's internal proof structure. DeepSeek-Prover-V2-7B has the strongest curve over most of the range, which means it is the most likely model to cover a large fraction of the subtheorems belonging to a given theorem. Goedel-Prover-V2-8B is competitive at very low thresholds, indicating that it often makes some initial progress, but it falls below DeepSeek at medium and high thresholds, suggesting lower probability of approaching full completion. Kimina-Prover-Distill-8B remains behind these two stronger models, while Goedel-Prover-SFT is weakest throughout most of the range. Thus, DeepSeek is not only the strongest at individual subtheorem solving, but also the most reliable at advancing substantially through an entire theorem group.

\begin{figure}
    \centering
    \includegraphics[width=0.99\linewidth]{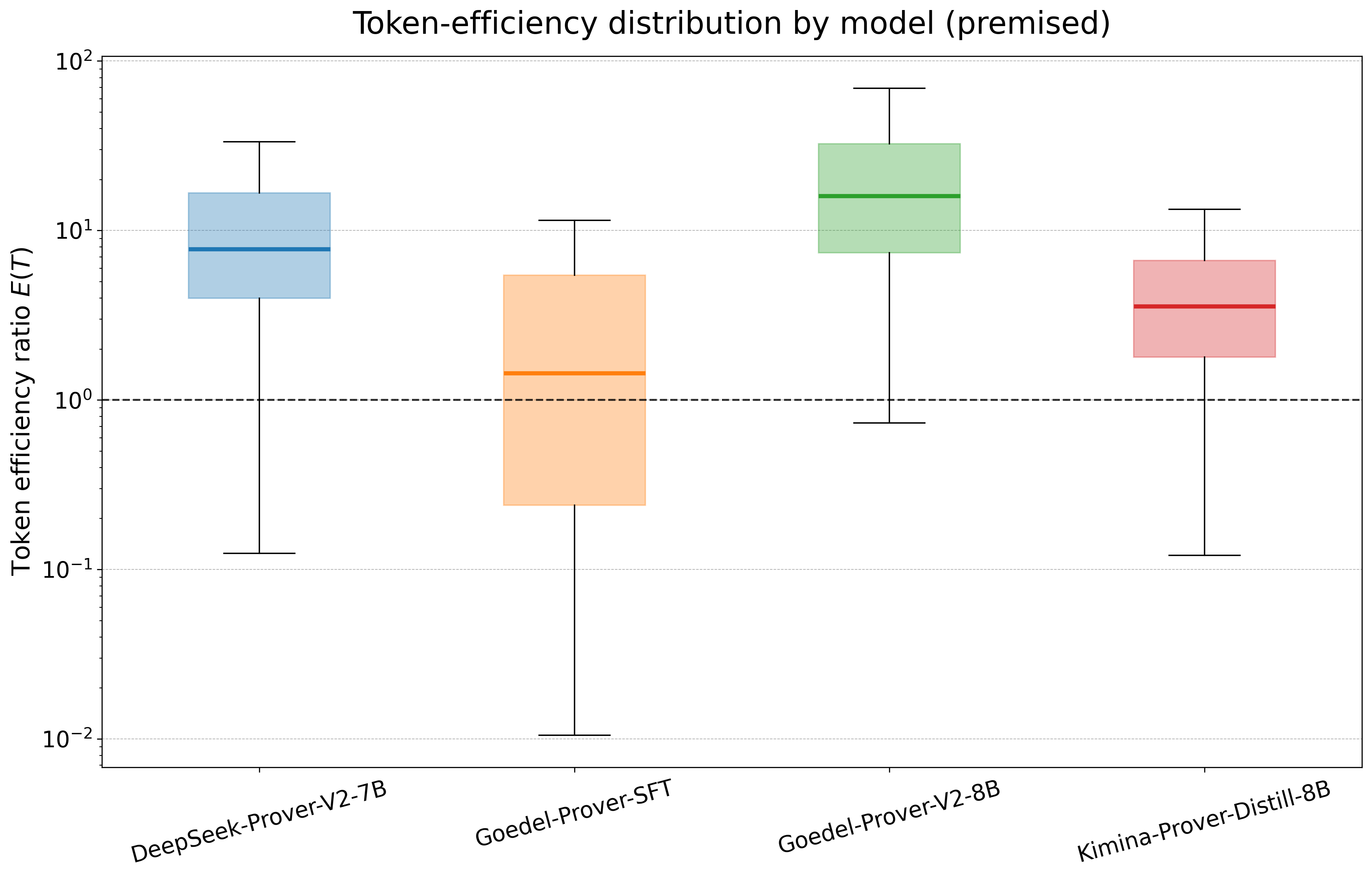}
    \caption{Overall token-efficiency distribution on solved premised instances. Values above $1$ indicate that the model's successful proofs are longer than the corresponding reference proofs.}
    \label{fig:token_efficiency_boxplot}
\end{figure}

\paragraph{Token-efficiency and proof verbosity.}
To compare generated proofs with reference proofs, we compute the token-efficiency ratio
\[
E(T)=\frac{\text{average successful generated proof tokens for }T}{\text{ground-truth proof tokens for }T}.
\]
Before tokenization, we remove non-Lean explanatory prose and comments, so the metric reflects proof text rather than natural-language reasoning. Figure~\ref{fig:token_efficiency_boxplot} shows that successful generated proofs are usually longer than the corresponding reference proofs. Goedel-Prover-V2-8B is the most verbose model, with median token-efficiency around $16$, followed by DeepSeek-Prover-V2-7B at around $7.8$. Kimina-Prover-Distill-8B is more compact, with median around $3.6$. Goedel-Prover-SFT has the lowest median, about $1.44$, but this reflects a much smaller and more selective solved subset rather than uniformly better proof compactness.

\paragraph{Discussion.}
Taken together, the results give a consistent picture of current Lean4 theorem-proving behavior. First, DeepSeek-Prover-V2-7B is the strongest overall model on TheoremBench under the premised setting, both in subtheorem pass@k and in theorem-level completion. Second, the premised representation is genuinely helpful, but only models that can use explicit auxiliary structure effectively obtain substantial gains from it. Third, performance is strongly biased toward easy subtheorems and toward a subset of mathematical fields, which shows that broad formal mathematical competence is still far from achieved. Finally, even when models succeed, their proofs are usually much longer than the corresponding ground-truth proofs, suggesting that correctness is often reached through verbose tactic exploration rather than compact proof planning. TheoremBench therefore exposes not only whether a theorem prover succeeds, but also how it succeeds, how far it progresses within a structured proof, and how efficiently it uses the formal context it is given.

\section{Limitations}

TheoremBench has several limitations. First, it is built from existing Lean4 formalizations of classical theorems, so its coverage depends on what has already been formalized and can be extracted into compilable snippets. Second, the premised transformation relies on syntactic dependency analysis and Lean-level verification; although every retained instance compiles, premise extraction may not always recover the mathematically minimal dependency set. Third, our experiments cover four Lean4-capable prover models under a fixed sampling budget and low model-size, leaving broader model families, search strategies, and retrieval-augmented settings for future work.

\section{Conclusion}
We introduced TheoremBench, a Lean4 benchmark for evaluating theorem provers on classical mathematical theorems beyond compact competition-style tasks. The benchmark provides two aligned settings: a plain-main version that tests final-statement proving, and a premised version that exposes supporting subtheorems and prior results as explicit premises. This structure enables more diagnostic evaluation, measuring not only whether a model proves a final theorem, but also how much of the underlying proof structure it can recover.

Our experiments show that explicit premises substantially help models able to exploit formal context, especially DeepSeek-Prover-V2-7B and Goedel-Prover-V2-8B, while weaker or non-reasoning models benefit little. The theorem-level coverage and token-efficiency analyses further show that current provers remain biased toward easy subtheorems and often produce verbose proofs. These results suggest that future Lean4 theorem provers should be evaluated not only by pass@k, but also by their ability to use structured premises, complete proof chains, and generate compact formal proofs.
%
%
%
\bibliographystyle{splncs04}
\bibliography{aist2026}

%
%
\newpage
\appendix
\section{Appendix}
\subsection{Dataset Distribution}
In total, TheoremBench contains 1,142 Lean4 proving instances across the processed theorem groups. Table \ref{tab:premised_dataset_distribution} summarizes the size of the extracted proof structures. Most theorem groups contain fewer than 20 supporting subtheorems, while a small number of long Lean developments contribute much larger groups.
\begin{table}
\centering
\begin{tabular}{c|c|c}
\toprule
\textbf{Subtheorems per theorem} & \textbf{Theorem groups} & \textbf{Premised instances} \\
\midrule
1 $-$ 5 & 39 & 117 \\
6 $-$ 10 & 9 & 81 \\
11 $-$ 20 & 19 & 319  \\
21 $-$ 50 & 14 & 460 \\
$>50$ & 2 & 165 \\
\midrule
\textbf{Total} & \textbf{83} & \textbf{1142} \\
\bottomrule
\end{tabular}
\caption{Distribution of supporting subtheorems per parent theorem in the retained premised dataset. A theorem group contains the main theorem and its extracted supporting subtheorems}
\label{tab:premised_dataset_distribution}
\end{table}
\subsection{Additional Results}
\begin{figure}
    \centering
    \includegraphics[width=1.0\linewidth]{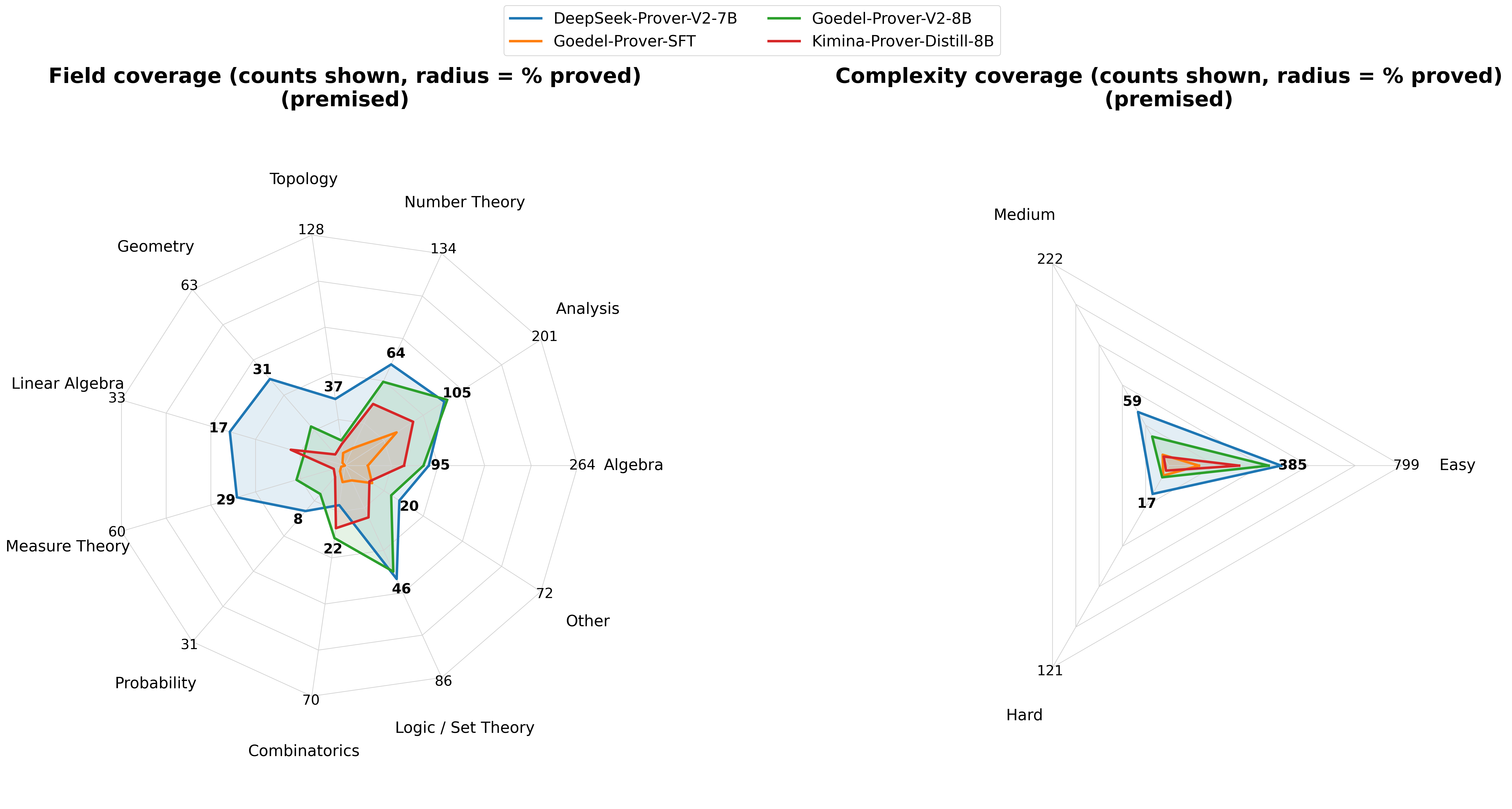}
    \caption{Coverage by mathematical field and complexity bucket on the premised benchmark. Counts are shown at vertices, while radius indicates percentage of proved instances. Performance is strongest on easy instances and concentrated in a subset of mathematical domains.}
    \label{fig:field_complexity_radar}
\end{figure}

\paragraph{Field and difficulty coverage.}
Figure~\ref{fig:field_complexity_radar} shows model coverage over mathematical fields and difficulty buckets. The field-level radar confirms that the benchmark spans a broad range of topics, but model performance is highly non-uniform across them. The strongest coverage is concentrated in algebra, analysis, number theory, and logic/set-theoretic material, while probability, measure theory, and several smaller fields remain much harder for all models. DeepSeek-Prover-V2-7B and Goedel-Prover-V2-8B have the broadest field coverage, although DeepSeek reaches the largest solved counts in the main categories. The complexity radar reveals an equally strong skew with respect to theorem difficulty. All models solve many more easy subtheorems than medium or hard ones. Even the strongest systems degrade sharply as the difficulty bucket increases. This confirms that current theorem provers remain much better at discharging short local obligations than at sustaining long and dependency-heavy proof chains.

\begin{figure}
    \centering
    \includegraphics[width=1.0\linewidth]{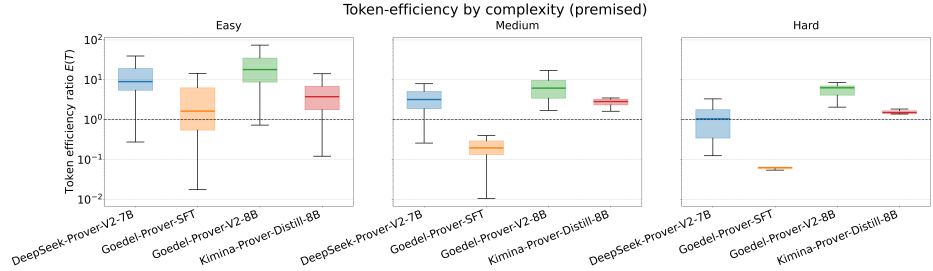}
    \caption{Token-efficiency conditioned on theorem difficulty. The strongest solved instances for several models become relatively more compact on harder buckets, but sample counts are much smaller there.}
    \label{fig:token_efficiency_by_complexity}
\end{figure}

\paragraph{Additional Token-Efficiency Analysis.}
Figure~\ref{fig:token_efficiency_by_complexity} conditions token-efficiency on theorem difficulty. For DeepSeek-Prover-V2-7B, the median token-efficiency drops from about $9.0$ on easy instances to about $3.2$ on medium instances and about $1.05$ on hard instances. Kimina-Prover-Distill-8B shows a similar trend, decreasing from about $3.8$ on easy instances to about $2.8$ on medium instances and about $1.5$ on hard instances. This suggests that, among the harder instances these models solve, generated proof lengths are closer to the references. Goedel-Prover-V2-8B remains comparatively verbose even on harder buckets, while Goedel-Prover-SFT has very small medians on medium and hard instances because it solves only a few selected examples. Table~\ref{tab:token_efficiency_appendix} reports the aggregate token-efficiency statistics used in the main text.

\begin{table}
\centering
\begin{tabular}{lcccc}
\toprule
Model & Solved instances & Median $E_{\mathrm{avg}}$ & Mean $E_{\mathrm{avg}}$ & Median $E_{\min}$ \\
\midrule
DeepSeek-Prover-V2-7B      & 529 & 7.78 & 20.72 & 4.69 \\
Goedel-Prover-SFT          & 127 & 1.44 & 9.65  & 0.20 \\
Goedel-Prover-V2-8B        & 381 & 16.00 & 42.49 & 8.51 \\
Kimina-Prover-Distill-8B   & 254 & 3.57 & 5.94  & 2.00 \\
\bottomrule
\end{tabular}
\caption{Token-efficiency summary on solved premised instances. Smaller values indicate more compact generated proofs relative to the reference Lean proofs, but values must be interpreted jointly with solved-instance counts.}
\label{tab:token_efficiency_appendix}
\end{table}

\begin{figure}
    \centering
    \includegraphics[width=0.99\linewidth]{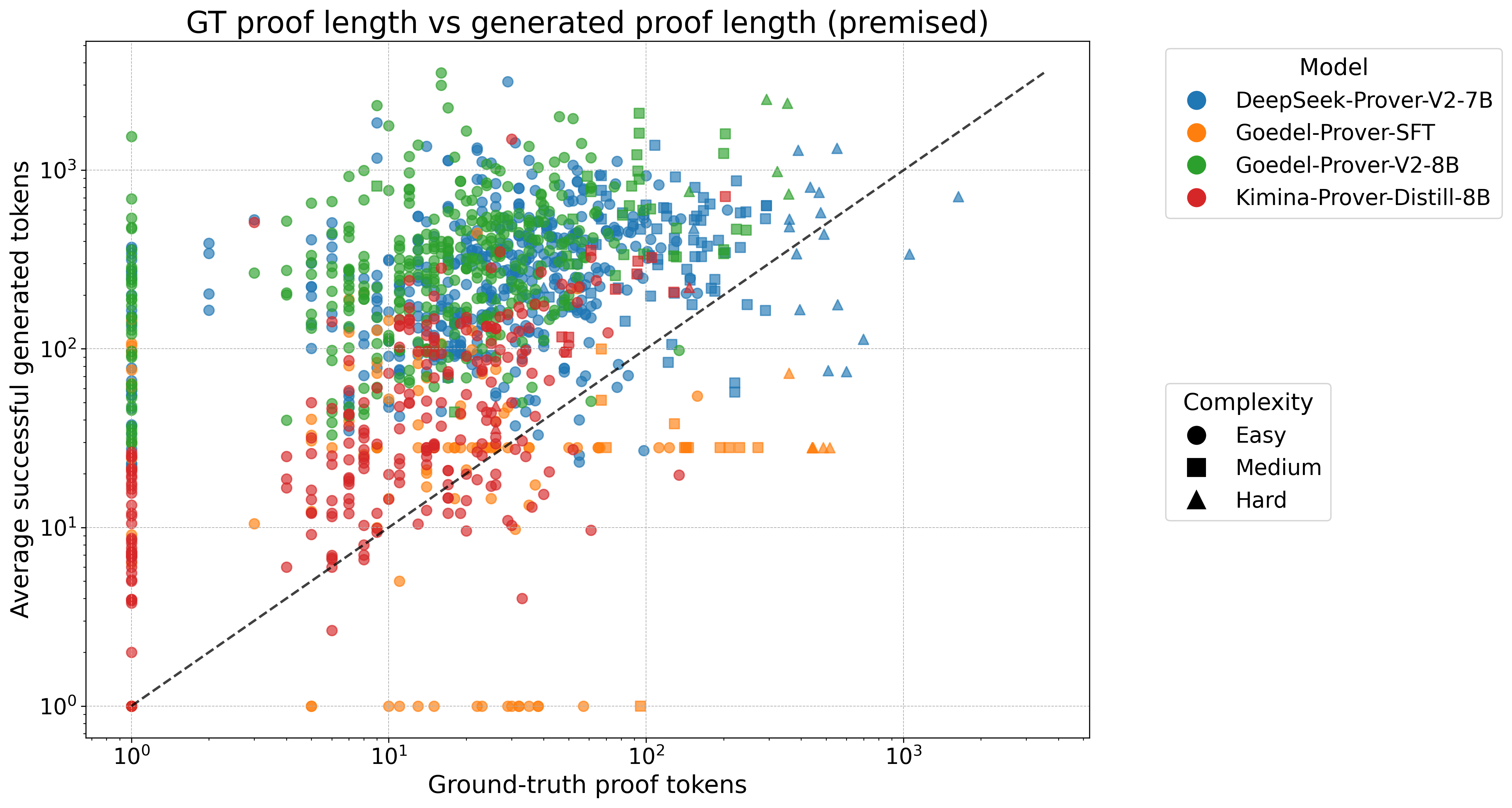}
    \caption{Ground-truth proof length versus average successful generated proof length. Most successful generations lie above the diagonal, showing that current provers typically produce longer proofs than the human-written references.}
    \label{fig:gt_vs_generated_scatter}
\end{figure}

\paragraph{Generated versus reference proof lengths.}
Figure~\ref{fig:gt_vs_generated_scatter} plots the token length of ground-truth proofs against the average length of successful generated proofs. The diagonal represents parity with the reference proof. Most DeepSeek and Goedel-Prover-V2-8B points lie well above the diagonal, indicating that these models often solve the theorem with substantially longer tactic traces than the human-written Lean proof. Kimina-Prover-Distill-8B lies closer to the diagonal on average, indicating more compact successful proofs. Goedel-Prover-SFT often lies closer to or even below the diagonal, but again only on a relatively small solved subset. The scatter therefore reinforces an important qualitative conclusion: current theorem provers are not simply memorizing or reproducing the reference proofs. Rather, they often reach correctness through much longer, more repetitive, and less economical proof scripts. The low token-efficiency of Goedel-Prover-SFT should be interpreted cautiously: it solves far fewer instances, so its compactness reflects a highly selective solved subset rather than uniformly better proof planning.

\paragraph{Pass@64 summary.}
Table~\ref{tab:pass64_summary_appendix} summarizes the final pass@64 values used in the main comparison plots.

\begin{table}
\centering
\begin{tabular}{lcc}
\toprule
Model & Premised pass@64 & Plain-main pass@64 \\
\midrule
DeepSeek-Prover-V2-7B      & 0.460 & 0.053 \\
Goedel-Prover-SFT          & 0.109 & 0.035 \\
Goedel-Prover-V2-8B        & 0.334 & 0.053 \\
Kimina-Prover-Distill-8B   & 0.223 & 0.053 \\
\bottomrule
\end{tabular}
\caption{Final pass@64 results on the two benchmark settings. The premised benchmark substantially improves performance for DeepSeek and Goedel-Prover-V2-8B, while gains are smaller for Kimina and negligible for Goedel-Prover-SFT.}
\label{tab:pass64_summary_appendix}
\end{table}

\subsection{Premises structure}

Figure~\ref{fig:examplepremisedstructure} illustrates how a supporting lemma is exposed as an explicit premise binder in the generated theorem statement.
\begin{figure}
\centering
\begin{examplebox}{Premises structure to original theorem statement}
\panelheading{Original Lean4 Statement}
\begin{codepanel}
\begin{leanverbatim}
import Mathlib.Algebra.Order.Archimedean.Basic
...

theorem card_eq_index_normalizer [Fact p.Prime] [Finite (Sylow p G)] (P : Sylow p G) :
    Nat.card (Sylow p G) = P.normalizer.index := by
    ...

theorem card_dvd_index [Fact p.Prime] [Finite (Sylow p G)] (P : Sylow p G) :
    Nat.card (Sylow p G) ∣ P.index := by sorry
\end{leanverbatim}
\end{codepanel}
\panelheading{Premised Lean4 Statement}
\begin{codepanel}
\begin{leanhlverbatim}
import Mathlib.Algebra.Order.Archimedean.Basic
...

theorem card_dvd_index_premised \premhl{\{card\_eq\_index\_normalizer :}
\premhl{([Fact p.Prime] \ensuremath{\to} [Finite (Sylow p G)] \ensuremath{\to} (P : Sylow p G) \ensuremath{\to}}
\premhl{Nat.card (Sylow p G) = P.normalizer.index\}} [Fact p.Prime]
[Finite (Sylow p G)] (P : Sylow p G) :
    Nat.card (Sylow p G) ∣ P.index := by sorry
\end{leanhlverbatim}
\end{codepanel}
\end{examplebox}
\caption{Example of transformation from original to premised structure. The highlighted binder exposes a previously proved supporting lemma as an explicit premise.}
\label{fig:examplepremisedstructure}
\end{figure}

\end{document}